\documentclass[10pt,twocolumn,letterpaper]{article}

\usepackage[pagenumbers]{cvpr} 

\usepackage{graphicx}
\usepackage{amsmath}
\usepackage{amssymb}
\usepackage{booktabs}
\usepackage{graphics} 
\usepackage{epsfig} 
\usepackage{mathptmx} 
\usepackage{times} 
\usepackage{amsmath} 
\usepackage{amssymb}  
\usepackage{gensymb}

\usepackage{cite}

\usepackage{ragged2e}
\usepackage{soul}
\usepackage{empheq}

\usepackage[dvipsnames]{xcolor}

\usepackage{graphicx}
\usepackage{amsmath}
\usepackage{amssymb}
\usepackage{booktabs}

\usepackage{graphics} 
\usepackage{mathptmx} 
\DeclareMathAlphabet{\mathcal}{OMS}{cmsy}{m}{n}
\usepackage{cite}

\usepackage{times}
\usepackage{epsfig}
\usepackage{graphicx}
\usepackage{amsmath}
\usepackage{amssymb}
\usepackage{color, colortbl}

\usepackage{helvet}
\usepackage{courier}

\usepackage{url}
\usepackage{rotating}
\usepackage{xspace}
\usepackage{booktabs}       
\usepackage{nccmath}
\usepackage{amsfonts}       
\usepackage{nicefrac}       
\usepackage{microtype}      
\usepackage{comment}		
\usepackage{adjustbox}		
\usepackage{tabu}
\usepackage{calc}			
\usepackage{wrapfig}

\usepackage{multirow}
\usepackage{float}
\usepackage[hang,flushmargin]{footmisc}
\usepackage[neverdecrease]{paralist}
\usepackage{caption}
\usepackage{subcaption}
\usepackage{algorithm}
\usepackage{algorithmic}
\usepackage{calc}
\usepackage{color}
\usepackage[font=footnotesize]{caption}
\usepackage{pifont}
\usepackage{makecell}
\usepackage{array}

\makeatletter
\@namedef{ver@everyshi.sty}{}
\makeatother
\usepackage{tikz}

\definecolor{LightCyan}{rgb}{0.88,1,1}
\definecolor{White}{rgb}{1,1,1}
\definecolor{CuGray}{gray}{0.95}

\newcommand{\figref}[1]{Fig.~\ref{#1}}
\newcommand{\tabref}[1]{Table~\ref{#1}}
\newcommand{\eqnref}[1]{Eq.~(\ref{#1})}
\newcommand{\secref}[1]{Sec.~\ref{#1}}

\makeatletter
\DeclareRobustCommand\onedot{\futurelet\@let@token\@onedot}
\def\@onedot{\ifx\@let@token.\else.\null\fi\xspace}

\def\etc{\emph{etc}\onedot} 
 
\def\etal{\emph{et al}\onedot}

\usepackage{mathtools}
\DeclarePairedDelimiter\abs{\lvert}{\rvert}%
\usepackage[pagebackref,breaklinks,colorlinks]{hyperref}
\usepackage[accsupp]{axessibility}

\usepackage[capitalize]{cleveref}
\crefname{section}{Sec.}{Secs.}
\Crefname{section}{Section}{Sections}
\Crefname{table}{Table}{Tables}
\crefname{table}{Tab.}{Tabs.}

\def\cvprPaperID{11014} 
\def\confName{CVPR}
\def\confYear{2023}

\begin{document}

\title{DualRefine: Self-Supervised Depth and Pose Estimation \linebreak Through Iterative Epipolar Sampling and Refinement Toward Equilibrium}



\author{Antyanta Bangunharcana$^{1}$, Ahmed Magd$^{2}$, Kyung-Soo Kim$^{1}$ \\
$^{1}$Mechatronics, Systems, and Control Laboratory, $^{2}$Vehicular Systems Design and Control Lab \\
Korea Advanced Institute of Science and Technology (KAIST), Republic of Korea \\
{\tt\small \big\{antabangun, a.magd, kyungsoo\}@kaist.ac.kr}
}

\maketitle

\begin{abstract}
    

    Self-supervised multi-frame depth estimation achieves high accuracy by computing matching costs of pixel correspondences between adjacent frames, injecting geometric information into the network. These pixel-correspondence candidates are computed based on the relative pose estimates between the frames. Accurate pose predictions are essential for precise matching cost computation as they influence the epipolar geometry. Furthermore, improved depth estimates can, in turn, be used to align pose estimates.

Inspired by traditional structure-from-motion (SfM) principles, we propose the DualRefine model, which tightly couples depth and pose estimation through a feedback loop. Our novel update pipeline uses a deep equilibrium model framework to iteratively refine depth estimates and a hidden state of feature maps by computing local matching costs based on epipolar geometry. Importantly, we used the refined depth estimates and feature maps to compute pose updates at each step. This update in the pose estimates slowly alters the epipolar geometry during the refinement process. Experimental results on the KITTI dataset demonstrate competitive depth prediction and odometry prediction performance surpassing published self-supervised baselines.\footnote{\url{https://github.com/antabangun/DualRefine}}

\end{abstract}

\section{Introduction}

\begin{figure}[ht!]
\begin{center}
    \begin{adjustbox}{width=0.25\textwidth}
    \setlength{\tabcolsep}{1pt}
    \begin{tabular}{c}
    \includegraphics[width=0.5\textwidth]{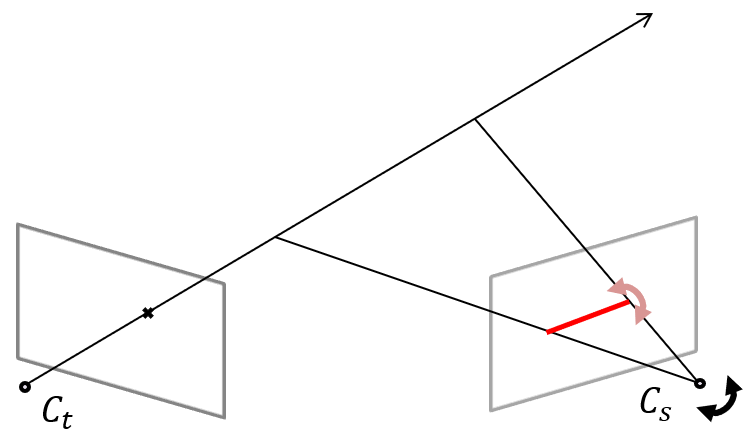} \\
    (a) \\
    \\
    \includegraphics[width=0.5\textwidth]{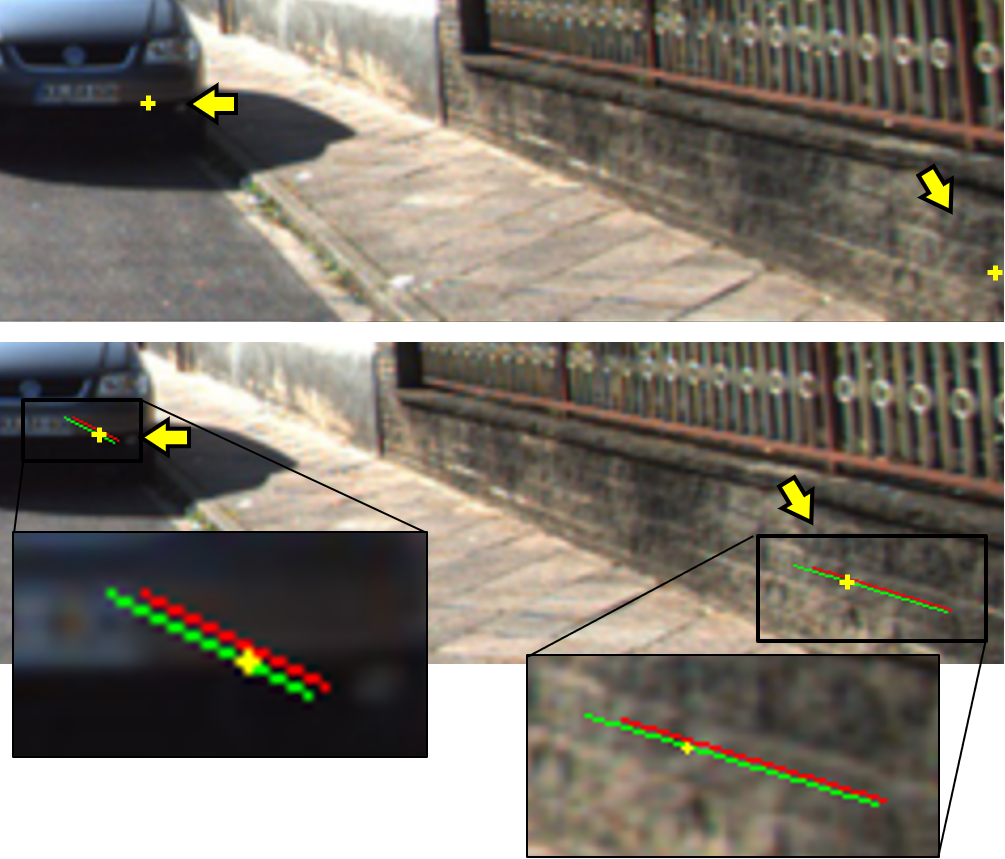} \\
    (b)
    \end{tabular}
    \end{adjustbox}
\caption{(a) The estimated pose of a camera affects the epipolar geometry. (b) The epipolar line in the source image, calculated from \textcolor{yellow}{yellow} points in the target image, for the PoseNet-based~\cite{kendall2015posenet} initial pose regression (\textcolor{red}{red}) and our refined pose (\textcolor{green}{green}). The yellow point in the source image is calculated based on our final depth and pose estimates.}
\vspace{-7mm}
\label{fig:epi}
\end{center}
\end{figure}

The optimization of the coordinates of observed 3D points and camera poses forms the basis of structure-from-motion (SfM). Estimation of both lays the foundation for robotics~\cite{hong2021visual, hong2022robotic, shyam2020retaining}, autonomous driving~\cite{geiger2012we}, or AR/VR applications~\cite{luo2020consistent}. Traditionally, however, SfM techniques are susceptible to errors in scenes with texture-less regions, dynamic objects, \etc. This has motivated the development of deep learning models that can learn to predict depth from monocular images~\cite{eigen2014depth, eigen2015predicting, li2015depth, laina2016deeper, fu2018deep}. These models can accurately predict depth based solely on image cues, without requiring geometric information.

In recent years, self-supervised training of depth and pose models has become an attractive method, as it alleviates the need for ground truth while demonstrating precision comparable to those of supervised counterparts~\cite{garg2016unsupervised, godard2017unsupervised, zhou2017unsupervised, mahjourian2018unsupervised, yin2018geonet, wang2018learning, zou2018df, bian2019unsupervised, shu2020feature, guizilini2020semantically, ranjan2019competitive, godard2019digging, gordon2019depth, guizilini20203d, vasiljevic2020neural}. Such an approach uses depth and pose predictions to synthesize neighboring images in a video sequence and enforce consistency between them. As the image sequence is also available at test time, recent self-supervised methods also study the use of multiple frames during inference~\cite{watson2021temporal}. These typically involve the construction of cost volumes from multiple views to compute pixel correspondences, bearing similarities to (multi-view) stereo models~\cite{kendall2017end, teed2018deepv2d, bangunharcana2021correlate}. By incorporating multi-frame data, geometric information is integrated to make depth predictions, improving the performance as well as the robustness. In such a multi-frame matching-based model, the accuracy of matching costs computation is essential. Recent work in DepthFormer~\cite{guizilini2022multi} demonstrates its importance, as they designed a Transformer~\cite{vaswani2017attention}-based module to improve matching costs and achieve state-of-the-art (SoTA) depth accuracy. However, their approach came with a large memory cost.


Unlike stereo tasks, the aforementioned self-supervised multi-frame models do not assume known camera poses and use estimates learned by a teacher network, typically a PoseNet~\cite{kendall2015posenet}-based model. This network takes two images as input and regresses a 6-DoF pose prediction. 
As the estimated pose affects the computation of epipolar geometry (\figref{fig:epi}(a)), the accuracy of the pose estimates is crucial to obtain accurate correspondence matches between multiple frames. However, as noted in recent studies~\cite{sarlin2021back}, pure learning-based pose regression generally still lags behind its traditional counterpart, due to the lack of geometric reasoning. By refining the pose estimates, we can improve the accuracy of the matching costs, potentially leading to better depth estimates as well. In \figref{fig:epi}(b), we show that the epipolar lines calculated from the regressed poses do not align with our refined estimates.
Conversely, a better depth prediction may lead to a better pose prediction. Thus, instead of building the cost volume once using regressed poses, we choose to perform refinements of both depth and pose in parallel and sample updated local cost volumes at each iteration. This approach is fundamentally inspired by traditional SfM optimization and is closely aligned with feedback-based models that directly couple depth and pose predictions~\cite{gu2021dro}.

In this work, we propose a depth and pose refinement model that drives both towards an equilibrium, trained in a self-supervised framework. We accomplish this by making the following contributions:
\textbf{First}, We introduce an iterative update module that is based on epipolar geometry and direct alignment. We sample candidate matches along the epipolar line that evolves based on the current pose estimates. Then the sampled matching costs are used to infer per-pixel confidences that are used to compute depth refinements.
The updated depth estimates are then used in direct feature-metric alignments to refine the pose updates towards convergence. As a result, our model can perform geometrically consistent depth and pose updates. 
\textbf{Second}, These updates refine the initial estimates made by the single-frame model. By doing so, we do not rely on full cost volume construction and base our updates only on local cost volumes, making it simpler, more memory efficient, and more robust. 
\textbf{Lastly}, we design our method within a deep equilibrium (DEQ) framework~\cite{bai2019deep} to implicitly drive the predictions towards a fixed point. Importantly, DEQ allows for efficient training with low training memory, improving upon the huge memory consumption of previous work. 
With our proposed novel design, we show improved depth estimates through experiments that are competitive with the SoTA models. Furthermore, our model demonstrates improved global consistency of visual odometry results, outperforming other learning-based models.

\label{sec:introduction}

\section{Related Work}

\subsection{Depth from a single image}


The depth prediction problem for a single image is ill-posed due to the possibility of different 3D scenes projecting onto the same 2D image \cite{hartley2003multiple}.
Nonetheless, humans can predict depth from a single image through experience. Motivated by this, numerous supervised neural network models were proposed to solve the monocular depth estimation task, starting with Eigen \etal.'s paper \cite{eigen2014depth}. Subsequently, performance was improved by modifying the model architecture \cite{eigen2015predicting, laina2016deeper, mousavian2016joint, wofk2019fastdepth}, training on large dataset \cite{Chen_2020_CVPR, xian2018monocular, li2018megadepth, zamir2018taskonomy}, designing robust loss functions \cite{lee2019big, zhang2018joint}, and transforming the problem into a classification task \cite{li2018deep}. However, supervisory depth estimation requires ground-truth depth maps, which are difficult to collect in large quantities and of high quality. This challenge is one of the main reasons why researchers are exploring semi-supervised training, where the model expects weak supervision, such as providing relative depth \cite{chen2016single}, camera poses \cite{zhan2018unsupervised}, or utilizing synthetic data for training \cite{atapour2018real, kundu2018adadepth, mayer2018makes}.

The need for weak supervision still presents limitations in generalizability and scalability, among other aspects. To address these constraints, research on self-supervised training techniques is gaining momentum. These techniques involve using geometry in stereo matching \cite{garg2016unsupervised, godard2017unsupervised} or with a sequence of single-camera images, as initially proposed by \cite{zhou2017unsupervised}. Monodepth2 \cite{godard2019digging} refined the idea of exploiting image sequences for training by using auto-masking and minimum reprojection losses to address occlusion and ego-motion issues. 
Further improvements were made by defining the problem as a classification task \cite{gonzalezbello2020forget, johnston2020self}, modifying the architecture \cite{guizilini20203d}, feature-based loss for regions with low texture~\cite{shu2020feature, zhan2018unsupervised}, or reducing artifacts from moving objects \cite{casser2019unsupervised, klingner2020self, tosi2020distilled}. 

Our work is based on self-supervised monocular depth, which is employed as a teacher and initial estimate.

\subsection{Depth from multiple frames}
Relying on single frames at test-time requires the model to make several assumptions about the scene's geometrical details. In contrast, multi-frame approaches, which leverage available temporal information and incorporate multi-view geometry, reduce the need for such assumptions.

Multi-frame depth prediction is closely related to stereo depth estimation, where neural networks convert input stereo images into depth maps, as demonstrated by \cite{mayer2016large, liang2018learning, ummenhofer2017demon}. Kendall \etal \cite{kendall2017end} achieved a significant improvement by constructing a plane-sweep stereo cost volume. Generally, multi-view stereo (MVS) research is more relevant to our work, as it utilizes an unstructured collection of scene images, meaning that the pose between different images is not fixed. In studies like \cite{huang2018deepmvs, im2019dpsnet, long2020occlusion, wang2018mvdepthnet, yao2018mvsnet}, it was common to combine the previously mentioned cost volumes with ground truth depth and camera poses for guidance. Consequently, these works require camera poses during inference \cite{wei2020deepsfm, liu2019neural}. Although some research managed to relax the pose requirement during inference, training still necessitates this supervision \cite{ummenhofer2017demon, teed2018deepv2d}.

The methods discussed above, along with advances in other domains such as scale domain adaptation \cite{guizilini2021geometric, zhao2019geometry, pnvr2020sharingan}, view synthesis \cite{godard2019digging, wei2021nerfingmvs}, and other supervisory techniques \cite{gordon2019depth, guizilini2020semantically}, have significantly impacted the performance of self-supervised approaches. ManyDepth \cite{watson2021temporal}, which is similar to our work, focuses on the drawbacks of using a single image during inference and proposes a flexible model that takes advantage of multiple frames at test time, if available. ManyDepth reduces artifacts from moving objects and temporarily stationary cameras by using a single-frame model as a teacher, resulting in improved depth map accuracy. More recently, DepthFormer~\cite{guizilini2022multi} achieved substantial accuracy improvements by using Transformers~\cite{vaswani2017attention} to obtain improved pixel matching costs.

Inspired by traditional bundle adjustments, we design a model that simultaneously solves for both depth and pose while incorporating many of the advancements mentioned earlier. Our approach is most similar to DRO~\cite{gu2021dro}. However, compared to their approach, our model tightly integrates multi-view geometry into the iterative updates formulation and bases our refinements on the local epipolar geometry.

\subsection{Iterative refinements}
Iterative refinement has been used to improve prediction quality in various learning tasks, including object detection \cite{bardhan2020salient, gong2019improving}, optical flow estimation \cite{hur2019iterative, teed2020raft}, semantic segmentation \cite{pinheiro2016learning, zhang2019canet}, and others \cite{jia2022segment, saharia2021image}. Some recent research has attempted to iterate the refinement process using deep convolutional networks \cite{casanova2018iterative, ghiasi2016laplacian, lin2017refinenet, liu2021confidence}. Other works train the same network repeatedly by utilizing the results of the previous iteration \cite{yu2018recurrent, zhang2019canet}. 
In particular, RAFT~\cite{teed2020raft} found success with its iterative refinement procedure for flow estimation. DEQ-flow~\cite{bai2022deep} employed a deep equilibrium (DEQ)~\cite{bai2019deep} framework to reduce the memory consumption of RAFT during training while maintaining accuracy.

A key component of our model is inspired by the iterative updates of RAFT and DEQ-flow. Instead of optical flow, our model refines depth and pose estimates in parallel. We design our refinement module to tightly couple the two predictions, considering the epipolar geometry of adjacent frames. With every update, the epipolar geometry is refined, which also results in a more accurate matching costs computation of pixel correspondences.

\subsection{Pose estimation}
Pose estimation is a crucial component in self-supervised monocular depth models. In many works, PoseNet-based models~\cite{kendall2015posenet} take a pair of adjacent image frames and output a 6 DoF pose estimation. This class of models is straightforward, but often less accurate than their traditional counterparts~\cite{mur2017orb, engel2014lsd, engel2017direct} due to the absence of geometry constraints. Recent work in deep learning-based localization has adopted differentiable geometrically inspired designs within their models by using direct alignments~\cite{von2020gn, von2020lm, sarlin2021back} or geometric alignments based on optical flow~\cite{teed2021droid}. These models demonstrate better generalization properties and accuracy. The use of self-supervised monocular depth has also been used to improve traditional odometry~\cite{yang2020d3vo, bian2021unsupervised}. However, our interest lies in learning to refine both depth and pose in the self-supervision pipeline.
To improve the accuracy of pose estimation, we integrate direct alignment within the recurrent module, ensuring geometrically consistent prediction between depth and pose.

\label{sec:related}

\section{Method}

\begin{figure*}[t!]
\begin{center}
    \includegraphics[width=1\textwidth,trim={0 46mm 0 32mm},clip]{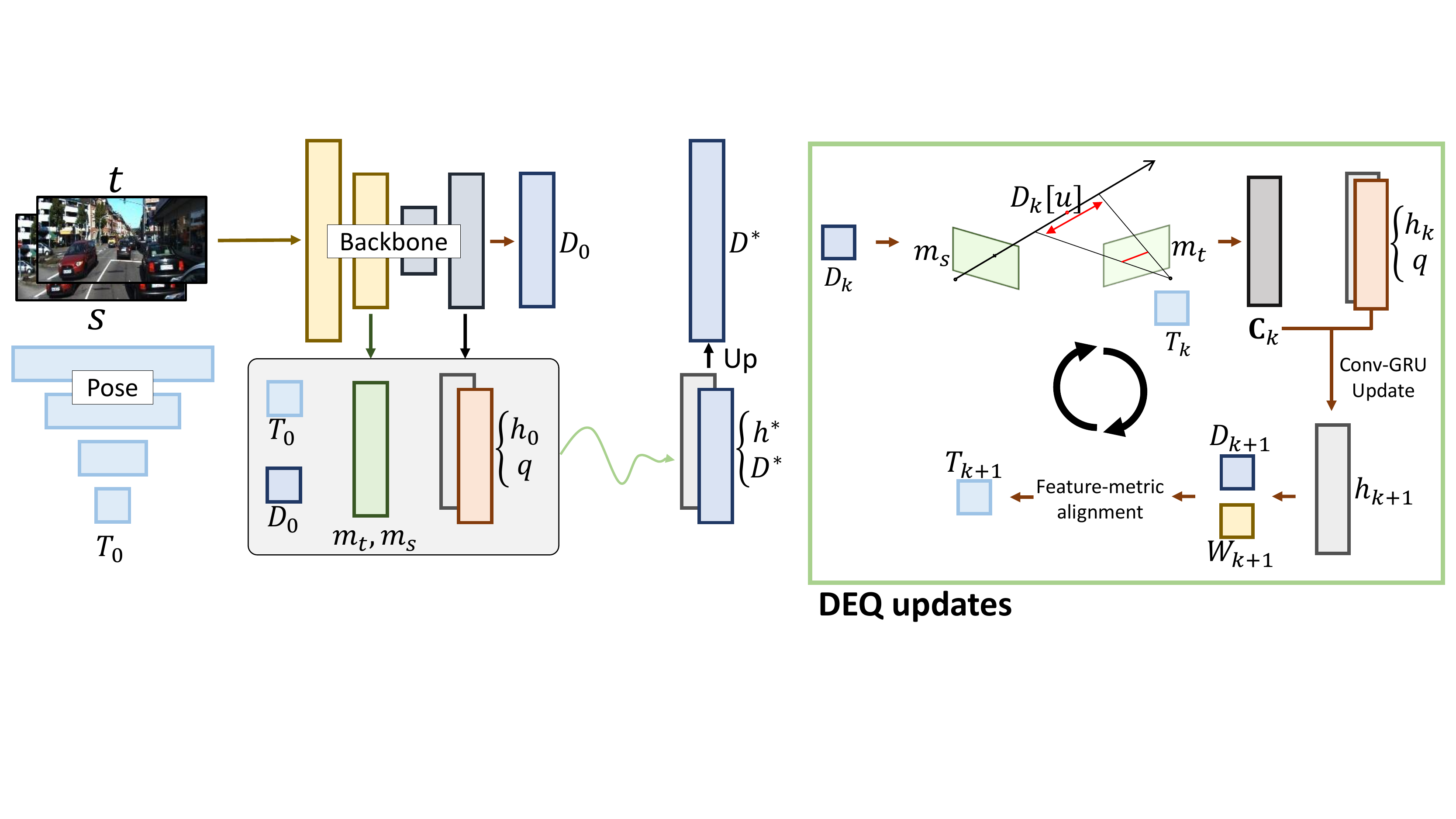}
\caption{(a) The overall pipeline of the model. Given a pair of source and target images, the teacher model predicts an initial depth $D_0$ and pose $T_0$, as well as initial hidden states that will be updated. DEQ-based alignments are then performed to find the fixed point and output the final predictions. (b) Each iteration in the update step takes the current depth and pose estimates. Matching costs are sampled along the current epipolar lines that evolves based on the pose estimates. The updates are computed by Conv-GRU. Then feature-metric alignment is used to obtain a geometrically consistent pose update. }
\vspace{-7mm}
\label{fig:overall}
\end{center}
\end{figure*}

Our model comprises two primary sub-modules. The first is a single frame self-supervised depth and pose estimator, building upon previous frameworks~\cite{godard2019digging, zhou2021self}, which we revisit in \secref{sec:monodepth}. This network serves as both a teacher and an initializer for the second sub-module, our proposed multi-frame network, presented in detail in \secref{sec:iter}.

\subsection{Self-supervised depth and pose}
\label{sec:monodepth}

We begin by describing the canonical self-supervised monocular depth estimation pipeline~\cite{godard2019digging}, which serves as the foundation for our approach. This depth training method assumes that a monocular camera with an intrinsic parameter $K$ captures an image sequence of a scene. In this process, two networks are trained in parallel to estimate the per-pixel depth map of the images $D$ and the relative poses between adjacent image frames. By warping neighboring images towards a shared target frame using these two predictions, self-supervised training can be performed by enforcing photometric consistency between the frames. 

Given the depth map $D$ of a target image and its relative pose with a source image $T_{t\rightarrow s}$, we can calculate the projection of each pixel $u=(x,y)$ of the target image onto the source image as follows:
\begin{equation}
    z'u'=z'\begin{bmatrix}x'\\ y'\\ 1\end{bmatrix} = KT_{t\rightarrow s}\Biggl(D[u]K^{-1}\begin{bmatrix}x\\ y\\ 1\end{bmatrix}\Biggr),
    \label{eq:proj}
\end{equation}
using the estimated depth at that pixel $D[u]$. The source images can then be warped towards the target frame by sampling the pixel values at the calculated projection
\begin{equation}
    I_{s\rightarrow t}[u] = I_{s}\langle u'\rangle,
    \label{eq:warp}
\end{equation}
where $\langle \cdot \rangle$ indicates bilinear interpolation, implemented using the spatial transformer network (STN)~\cite{jaderberg2015spatial}.


The self-supervised loss is calculated as a combination of the photometric error and the edge-aware smoothness loss:
\begin{equation}
    \mathcal{L}_{self-sup} = \lambda_p \mathcal{L}_p + \lambda_s \mathcal{L}_s,
    \label{eq:loss}
\end{equation}
where the photometric error between the warped source image and the target image is calculated using the structural similarity loss, and the minimum error is taken between multiple warped images to account for occlusions. Interested readers can refer to ~\cite{godard2019digging}.
The self-supervised loss is typically computed at multiple scales to stabilize the training.

In this paper, we train a monocular depth estimator and a pose estimation network using this pipeline, serving two purposes. First, following ManyDepth~\cite{watson2021temporal}, we use these models as a teacher to constrain the multi-frame predictions in the presence of dynamic objects. Additionally, we employ them as an initializer for the multi-frame alignment network.

\subsubsection{Monocular model}
We build our monocular depth estimation model based on DIFFNet ~\cite{zhou2021self}, a SoTA self-supervised single-frame estimator. We extract feature maps at multiple scales $s$ from the target image using the HRNet architecture~\cite{wang2020deep}.  In accordance with DIFFNet, feature maps from multiple stages are accumulated in $F^{(1/2^s)}$. Then, we employ disparity decoders to make disparity predictions at scales $s=\{2,3\}$. The pose estimation network follows the canonical PoseNet~\cite{kendall2015posenet} architecture, taking two input images and outputting 6-DoF values, with a ResNet18~\cite{he2016deep} backbone. The predicted disparity and pose estimation from these networks are used as a teacher to train our alignment sub-module, as was done in ManyDepth~\cite{watson2021temporal}.

\subsection{Deep equilibrium alignments}
\label{sec:iter}

In our alignment sub-module, we assume that additional input from source image(s) is available, which can be used to refine the depth and pose estimates. In this work, we focus on using the image from the previous frame in the image sequence as our source image.

Our alignment module is formulated as a deep equilibrium model~\cite{bai2019deep} that updates the hidden states, depth, and pose estimates to a fixed point. Specifically, at the fixed point $z^*$, 
\begin{equation}
\begin{split}
    (h^*, D^*, T^*) = z^* = \mathrm{U}(z^*, x),
\end{split}
\end{equation}
where $z$ is composed of a hidden state $h$, the depth prediction $D$, and the pose prediction $T$. $\mathrm{U}$ represents our update function, refining the depth and pose alternatively. $x$ is an input to the update module, obtained based on the epipolar geometry at each step, which we discuss in the next subsection. We perform these iterative updates at scale $s=2$.

From the feature extraction output, we compute an initial hidden state for our recurrent updates $h^{[0]}=\tanh(\mathrm{H}(F^{(1/4)}))$ and a context feature $q=\mathrm{Q}(F^{(1/4)})$. Both $\mathrm{H}$ and $\mathrm{Q}$ are composed of a single residual block~\cite{he2016deep} followed by a convolutional layer. 

\begin{figure*}[t!]
\begin{center}
    \begin{adjustbox}{width=1\textwidth}
    \setlength{\tabcolsep}{1pt}
    \begin{tabular}{c}
    \includegraphics[width=1\textwidth]{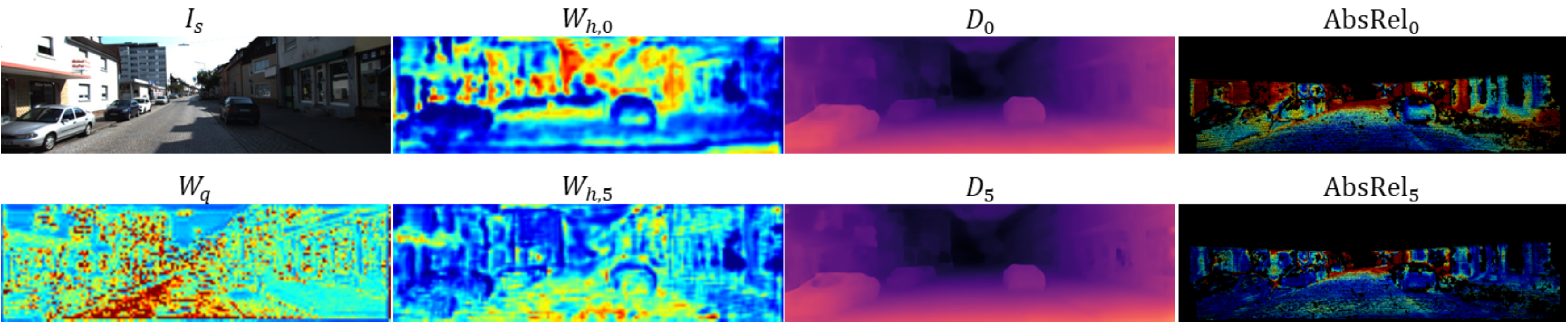} \\
    \hline
    \includegraphics[width=1\textwidth]{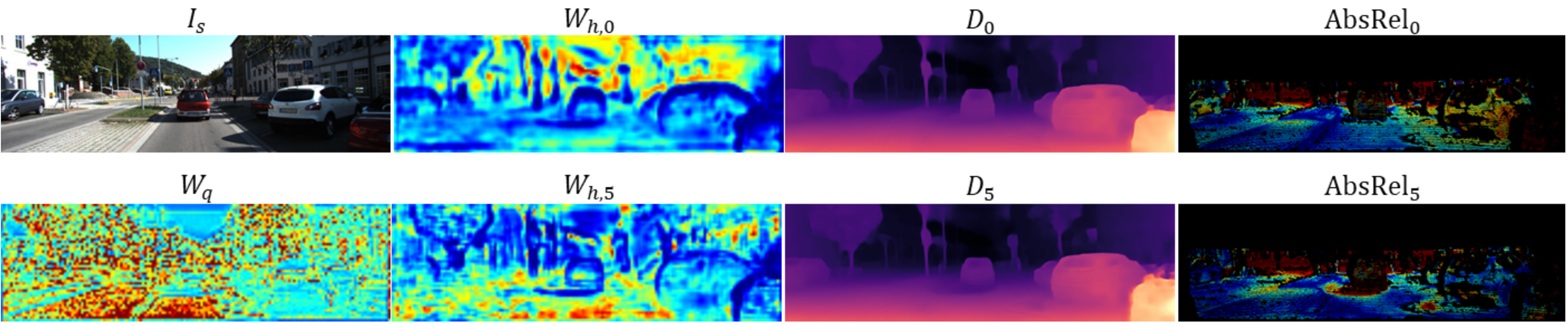} \\
    \end{tabular}
    \end{adjustbox}
\caption{Qualitative results on KITTI data. $I_s$: input image; $W_q$, $W_{h,0}$, and $W_{h,5}$: confidence weights; $D_0$, $D_5$: disparity estimates; The Abs Rel error for the depth estimates.}
\vspace{-7mm}
\label{fig:res1}
\end{center}
\end{figure*}

\subsubsection{Depth updates around local neighborhood}
\label{seq:matching}

\noindent\textbf{Local epipolar sampling}. As discussed previously, our refinement is based on the pixel matching between the target image and the source image. We use the first two blocks of the HRNet feature extractor to extract unary features from the source image $m^{(1/4)}_s$ and the target image $m^{(1/4)}_t$, which we will use to calculate the matching costs.

Similarly to RAFT~\cite{teed2020raft}, our aim is to compute the matching values for candidate correspondences around the current prediction and use them as input to our update module at the update step $k$. Unlike RAFT, however, we perform matching along an epipolar line based on the pose estimate $T_k$. This is done by computing the projected coordinates $u'_k$ in the source image obtained from \eqref{eq:proj} at the depths of interest. The depth candidates are computed to be the neighborhood of the current prediction. Instead of computing the all-pair correlation as was done in RAFT, we compute feature matching on the fly. 

At each pixel $u$, we compute the local depth candidates as $D[u]_k \pm (i\times c\times n)$, with $i\in\mathbb{Z}$ and $i\leq r$. We set the sampling radius hyperparameter $r=8$ in our experiments. In depth estimation, the error typically grows with distance. To account for this, we define $c=D[u]/C$ as a function of depth to make the sampling range dependent on the depth, where we set $C$ as a trainable parameter. Following RAFT, to collect the matching information from a larger neighborhood, we sampled at multiple levels $n=\{1, 2, 3\}$. At each level, we bilinearly resize the matching feature map of the source image $m^{(1/4)}_s$ with a scale of $1/2^n$. Then, the matching features are sampled at the calculated corresponding set of coordinates $u'_k$, and the absolute differences with the target feature
\begin{equation}
    \mathbf{C}_k[u] = \abs{m_t[u] - m_s\langle u'_k\rangle}
\end{equation}
are calculated and gathered. This provides us with a map $\mathbf{C}_k$ that contains $n\times (2\times r + 1)$ matching cost values at the corresponding depth candidates.
We then encode these matching costs along with the depth using a two-layer convolutional neural network ($\mathrm{CNN}$) to compute the input for the update module
\begin{equation}
\begin{split}
    x_k = [\mathrm{CNN}_{\mathbf{C}}(\mathbf{C}_k), \mathrm{CNN}_{D_x}(D_k), q],
\end{split}
\label{eq:xk}
\end{equation}
where $[\cdot, \cdot]$ represents the concatenation. 



\vspace{3mm}
\noindent\textbf{Depth update}. The update function calculates an updated hidden state $h_{k+1}$ using the Conv-GRU block~\cite{teed2020raft, jiang2021learning, bangunharcana2022revisiting}.
$h_{k+1}$ is used to compute the depth updates. 
To stabilize training, we use the activation function $\tanh(\cdot)$ to bound the absolute update values for the depths to be within $r\cdot c$: 
\begin{equation}
    D_{k+1} = D_{k} + r\cdot c\cdot  \tanh(\mathrm{CNN}_{D_\mathrm{U}}(h_{k+1})).
\label{eq:upd}
\end{equation}
These updates are performed in an alternating fashion with the pose updates to reach the fixed point $D^*$. 
Using $h^*$, we compute a convex upsampling to obtain the final depth estimate at the input resolution.


\subsubsection{Feature-metric pose alignments}
In \eqref{eq:proj}, the accuracy of pose estimation affects the calculation of the coordinates of pixels. Hence, a refined pose estimate would also improve the reliability of the matching costs. To refine the pose while being geometrically consistent, we perform our pose updates based on direct feature alignments~\cite{engel2014lsd, engel2017direct, wang2018learning}. 
These updates $\delta_k$ can be calculated by solving $H_k\delta_k=b_k$, where 
\begin{equation}
\begin{split}
    H_k=\mathcal{J}_k^Tdiag(W_k)\mathcal{J}_k \mbox{ and } b_k=-\mathcal{J}_k^Tdiag(W_k)r_k \\
    r_k[u] = m_s\langle u'_k\rangle - m_t[u],
\end{split}
\label{eq:hessian}
\end{equation}
and $\mathcal{J}$ is the Jacobian with respect to the pose. 

To compute this pose update, one could assign uniform confidence to every pixel in the image.
However, in \eqnref{eq:hessian}, additional confidence weights per pixel $W$ can also be integrated. This is done for two reasons. First, the solution for the pose updates can be affected by dynamic objects as well as inaccurate feature alignments that may occur in region with repeated textures. To account for this, the confidence weighting of the input context feature map $W_q$ can be computed~\cite{sarlin2021back}.
Specifically, we computed a confidence map $W_{t,s}=1/(1+\mathrm{ReLU}(\mathrm{CNN}_{W_q}(F_{t,s})))$ for both target and source images. The source confidence is warped towards the target frame using \cref{eq:proj,eq:warp}. Finally, the confidence is computed as $W_q=W_t\cdot W_{s,warped}$. This confidence weight is only computed once to assign per-pixel confidence for the input images.

Second, the accuracy of pose updates would depend on the accuracy of the depth estimates. To obtain more accurate pose updates, we would like to assign more alignment confidence to the region with higher depth accuracy. We use a neural network to infer this confidence from the matching costs. Since the hidden states $h$ have a history of these matching costs, another confidence weights $W_{h,k}=\mathrm{CNN}_{W_h}(h_k)$ can be computed. Unlike the previous confidence map, this one is computed and evolves at every update step. By doing so, the network can use the depth predictions to guide the pose estimates towards convergence using the depth information. In our experiments, we investigate the use of each confidence weighting and combination of both $W_k=W_qW_{h,k}$.

Finally, we can compute the updated pose as $T_{k+1}=exp(\delta_k^\wedge)T_{k}$. These operations are designed to be differentiable to enable end-to-end training.

\newcommand{\sbullets}{\tikz\draw[black,fill=black] (0,0) circle (.4ex)}

\begin{table*}[ht!]
    \begin{center}
        \begin{adjustbox}{width=1\textwidth}
        \begin{tabular}{c|c|ccc|ccccccc}
            \Xhline{4\arrayrulewidth}
            &Method & Test frames & Semantics & $W\times H$ & \cellcolor{red!7} Abs Rel $\downarrow$ & \cellcolor{red!7} Sq Rel $\downarrow$ &  \cellcolor{red!7} RMSE $\downarrow$ & \cellcolor{red!7} RMSE log $\downarrow$ & \cellcolor{blue!7} $\delta_1$ $\uparrow$ & \cellcolor{blue!7} $\delta_2$ $\uparrow$ & \cellcolor{blue!7} $\delta_3$ $\uparrow$ \\
            \hline
            \multirow{16}{*}{ \rotatebox{90}{Low \& mid res}} &
            Ranjan \etal~\cite{ranjan2019competitive} & 1 & & $832\times 256$ & 0.148 & 1.149 & 5.464 & 0.226 & 0.815 & 0.935 & 0.973 \\
            
            &EPC++~\cite{luo2019every} & 1 & & $832\times 256$ & 0.141 & 1.029 & 5.350 & 0.216 & 0.816 & 0.941 & 0.976\\
            
            &Struct2depth (M)~\cite{casser2019depth} & 1 & \sbullets; & $416\times 128$ & 0.141 & 1.026 & 5.291 & 0.215 & 0.816 & 0.945 & 0.979\\
            
            &Videos in the wild~\cite{gordon2019depth} & 1 & \sbullets; & $416\times 128$ & 0.128 & 0.959 & 5.230 & 0.212 & 0.845 & 0.947 & 0.976\\
            
            &Guizilini \etal~\cite{guizilini2020semantically} & 1 & \sbullets; & $640\times 192$ & 0.102 & 0.698 & 4.381 & 0.178 & 0.896 & 0.964 & \textbf{0.984}\\
            
            &Johnston \etal~\cite{johnston2020self} & 1 & & $640\times 192$ & 0.106 & 0.861 & 4.699 & 0.185 & 0.889 & 0.962 & 0.982\\
            
            &Monodepth2~\cite{godard2019digging} & 1 & & $640\times 192$ & 0.115 & 0.903 & 4.863 & 0.193 & 0.877 & 0.959 & 0.981\\
            
            &Packnet-SFM~\cite{guizilini20203d} & 1 & & $640\times 192$ & 0.111 & 0.785 & 4.601 & 0.189 & 0.878 & 0.960 & 0.982\\
            
            &Li \etal~\cite{li2020unsupervised} & 1 & & $416\times 128$ & 0.130 & 0.950 & 5.138 & 0.209 & 0.843 & 0.948 & 0.978\\
            &DIFFNet~\cite{zhou2021self} & 1 & & $640\times 192$ & 0.102 & 0.764 & 4.483 & 0.180 & 0.896 & \underbar{0.965} & \underbar{0.983} \\
            &\cellcolor{green!7}
            \textbf{DualRefine-initial ($D_0$)} & 1 & & $640\times 192$ & 0.103 & 0.776 & 4.491 & 0.181 & 0.894 & \underbar{0.965} & \underbar{0.983} \\
            
            \arrayrulecolor{gray}\cline{2-12}
            &Patil \etal~\cite{patil2020don} & N$^\dagger$ & & $640\times 192$ & 0.111 & 0.821 & 4.650 & 0.187 & 0.883 & 0.961 & 0.982\\
            &Wang \etal~\cite{wang2020self} & 2 (-1, 0) & & $640\times 192$ & 0.106 & 0.799 & 4.662 & 0.187 & 0.889 & 0.961 & 0.982 \\
            
            &ManyDepth (MR)~\cite{watson2021temporal} & 2 (-1, 0) & & $640\times 192$ & 0.098 & 0.770 & 4.459 & 0.176 & 0.900 & \underbar{0.965} & \underbar{0.983} \\
            &DepthFormer~\cite{guizilini2022multi} & 2 (-1, 0) & & $640\times 192$ & \underbar{0.090} & \textbf{0.661} & \textbf{4.149} & \underbar{0.175} & \underbar{0.905} & \textbf{0.967} & \textbf{0.984} \\
            &\cellcolor{green!7}
            \textbf{DualRefine-refined ($D^*$)} & 2 (-1, 0) & & $640\times 192$ & \textbf{0.087} & \underbar{0.698} & \underbar{4.234} & \textbf{0.170} & \textbf{0.914} & \textbf{0.967} & \underbar{0.983} \\
            \Xhline{4\arrayrulewidth}
            \multirow{4}{*}{ \rotatebox{90}{High res}} &
            DRO~\cite{gu2021dro} & 2 (-1, 0) & & $960\times 320$ & 0.088 & 0.797 & 4.464 & 0.212 & 0.899 & 0.959 & 0.980 \\
            &Wang \etal~\cite{wang2020self} & 2 (-1, 0) & & $1024\times 320$ & 0.106 & 0.773 & 4.491 & 0.185 & 0.890 & 0.962 & 0.982 \\
            &ManyDepth (HR ResNet50)~\cite{watson2021temporal} & 2 (-1, 0) & & $1024\times 320$ & \underbar{0.091} & \underbar{0.694} & \underbar{4.245} & \underbar{0.171} & \underbar{0.911} &\underbar{ 0.968} & \underbar{0.983} \\
            &\cellcolor{green!7}
            \textbf{DualRefine-refined (HR) ($D^*$)} & 2 (-1, 0) & & $960\times 288$ & \textbf{0.087} & \textbf{0.674} & \textbf{4.130} & \textbf{0.167} & \textbf{0.915} & \textbf{0.969} & \textbf{0.984} \\
            \Xhline{4\arrayrulewidth}
        \end{tabular}
        \end{adjustbox}
    \caption{Results and comparison with other state-of-the-arts models on the KITTI~\cite{geiger2012we} Eigen split. \textbf{Bold}: Best, \underbar{Underscore}: Second best. $^\dagger:$ evaluated on whole sequences}
    \vspace{-7mm}
    \label{table:comparison}
    \end{center}
\end{table*}

\subsubsection{DEQ training}

We adopt a DEQ framework, wherein the above steps are repeated until the depth and pose values reach a fixed point, at which the update value is minimal. In our implementation, the fixed points of depth and pose are chosen separately, and it is possible for both to be selected from different update steps. Finally, the training gradient is computed using the chosen depth and pose fixed points. Operations prior to the fixed point do not require saving gradients in memory, which allows memory-efficient training.


As noted in \cref{eq:proj,eq:warp,eq:loss}, the self-supervision losses can be computed by warping the input images given a depth and pose prediction. At the fixed point, we compute two additional self-supervision losses for each of the refined depth $D^*$ and pose $T^*$ estimates. In contrast to prior work, where multi-frame depth estimates $D^*$ are paired with the teacher pose estimate $T_0$ to perform warping for loss computation, we pair $D^*$ with $T^*$ instead. Detailed experiments for this choice of loss pairings can be found in the supplementary.


In both losses, we also apply the consistency loss between $D_0$ and $D^*$, similar to ManyDepth~\cite{watson2021temporal}, to account for dynamic objects or occluded regions. Specifically, we extract coarse depth predictions from the raw feature matchings and mask regions where large disagreements occur, and enforce consistency with the teacher depth. Unlike ManyDepth, however, our method does not explicitly construct a cost volume. To obtain the coarse depth, we search for the lowest matching cost around the neighborhood of the teacher depth, similar to \secref{seq:matching}, but with a larger neighborhood range. This approach offers an additional advantage compared to the cost volume-based method, as we do not need to rely on an estimated minimum and maximum depth or know the scale of the estimates.
Moreover, since the computation of coarse depth depends on the accuracy of matching costs, it can be improved with more accurate pose estimates (\cref{table:ablpose}). 

\label{sec:meth}

\section{Experiments}


\subsection{Dataset and metrics}

For depth estimation experiments, we use the Eigen train/test split~\cite{eigen2014depth} from the KITTI dataset~\cite{geiger2012we}. To evaluate the estimated depths, we scale them by a scalar to match the scale of the ground truth. We employ standard depth evaluation metrics~\cite{eigen2014depth, eigen2015predicting}, including absolute and squared relative error (Abs Rel, Sq Rel), root mean square error (RMSE, RMSE log), and accuracy under threshold ($\delta_1$, $\delta_2$, and $\delta_3$), with a maximum depth set at $80m$. Lower values are better for the first four metrics, while higher values are better for the remaining three.

For visual odometry experiments, we use the KITTI odometry dataset and follow the same training and evaluation sequences (Seq. 00-08 for training and Seq. 09-10 for evaluation) as in previous work~\cite{zhou2017unsupervised}. Since our estimation relies on a monocular camera, the estimated trajectories are aligned with the ground truth using the 7 DoF Umeyama alignment~\cite{umeyama1991least}. We use standard odometry evaluation metrics such as translation ($t_{err}$) and rotation ($r_{err}$) error ~\cite{geiger2012we}, and absolute trajectory error (ATE) ~\cite{sturm2012benchmark}.

\begin{table}[t]
    \begin{center}
        \begin{adjustbox}{width=0.48\textwidth}
        \begin{tabular}{cc|ccccc}
            \Xhline{4\arrayrulewidth}
            Pose & Consistency & \multirow{2}{*}{Abs Rel} & \multirow{2}{*}{Sq Rel} & \multirow{2}{*}{RMSE} & \multirow{2}{*}{$\delta_1$} & \multirow{2}{*}{$\delta_2$} \\
            Updates & mask & & & \\
            \Xhline{4\arrayrulewidth}
            no update            & $T_0$ & 0.097 & 0.713 & 4.462 & 0.898 & 0.964 \\
            \hline
            no weights           & $T_0$ & 0.091 & 0.694 & 4.271 & 0.909 & \textbf{0.967} \\
            no $W_{h,k}$         & $T_0$ & 0.090 & 0.667 & 4.252 & 0.909 & \textbf{0.967} \\
            no $W_q$             & $T_0$ & 0.093 & 0.686 & 4.258 & 0.908 & \textbf{0.967} \\
            $W_q$ and $W_{h,k}$  & $T_0$ & 0.090 & 0.669 & 4.293 & 0.910 & \textbf{0.967} \\
            \hline
            no weights           & $T^*$ & 0.092 & 0.667 & 4.257 & 0.908 & \textbf{0.967} \\
            no $W_{h,k}$         & $T^*$ & 0.091 & \textbf{0.666} & 4.243 & 0.909 & \textbf{0.967} \\
            no $W_q$             & $T^*$ & 0.088 & 0.674 & 4.251 & 0.911 & 0.966 \\
            $W_q$ and $W_{h,k}$  & $T^*$ & \textbf{0.087} & 0.698 & \textbf{4.234} & \textbf{0.914} & \textbf{0.967} \\
            \Xhline{4\arrayrulewidth}
        \end{tabular}
        \end{adjustbox}
    \caption{Ablation experiment for the effect of pose updates on the KITTI~\cite{geiger2012we} Eigen split. \textbf{Bold}: Best.}
    \vspace{-7mm}
    \label{table:ablpose}
    \end{center}
\end{table}

\subsection{Implementation details}

We conduct our experiments using PyTorch~\cite{paszke2019pytorch} on an RTX 3090 GPU with a batch size of 12. Following~\cite{godard2019digging}, we apply color and flip augmentations and resize input images to a resolution of $640\times 192$. For our high resolution experiments, we resize the images to $960\times 288$. We train the entire network for 15 epochs with a learning rate of $10^{-3}$, at which point we freeze the teacher depth and pose models. We then continue to train the network with a learning rate of $10^{-4}$. Adam optimizer~\cite{kingma2014adam} is employed with $\beta_1=0.9$ and $\beta_2=0.999$.

As mentioned earlier, the depth backbone is based on the HRNet architecture~\cite{wang2020deep}. For the teacher pose model, we adhere to the standard design, using the first five layers of a ResNet18 initialized with ImageNet pre-trained weights as an encoder, followed by a decoder that outputs 6-DoF pose estimates. At test time, depth estimates are made using the current frame and the previous frame when available. When the previous frame is unavailable, we skip the refinement module and simply use the initial estimates.

\subsection{Ablation}

\noindent
\textbf{Pose updates.}
We analyze the effect of pose refinement toward depth and present the findings in \tabref{table:ablpose}. Our model that does not perform pose updates has the worst accuracy.

\noindent
\textbf{Evolving confidence weighting.}
We also show the impact of confidence weightings. 
Interestingly, similar performance can be observed for all models with pose updates, even when no weighting is used to guide the pose computation.

\noindent
\textbf{Consistency mask.}
Here we use the refined pose to extract the consistency mask for training. The overall results are slightly improved, suggesting that the improved pose estimates help to compute more accurate matching costs.

\noindent
\textbf{DEQ iteration.}
We investigate aspects of DEQ iterative updates and present the findings when we vary the number of iterations during training and at test time in \tabref{table:abldeq}. The results suggest that 6 iterations is sufficient to find the fixed point. We speculate that the initial estimate provides a reliable starting point and hence fast convergence. Nevertheless, further investigation into training stabilization of models with a larger number of iterations using DEQ techniques~\cite{bai2022deep} is warranted. On our machine, the $6$ update iterations increase the baseline model inference time from $37.90$ ms (which also includes initial pose regression) by $\sim31$ ms to a total of $68.49$ ms, running at almost $15$ fps. However, we only use PyTorch basic functions in our implementation, and additional code optimization is could be made.

\begin{table}[t]
    \begin{center}
        \begin{adjustbox}{width=0.4\textwidth}
        \setlength{\tabcolsep}{8pt}
        \begin{tabular}{c|ccccc}
            \Xhline{4\arrayrulewidth}
            DEQ & \multirow{2}{*}{Abs Rel} & \multirow{2}{*}{Sq Rel} & \multirow{2}{*}{RMSE} & \multirow{2}{*}{$\delta_1$} & Time \\
            \# iters & & & & & (ms) \\
            \Xhline{4\arrayrulewidth}

            3$\rightarrow$3   & 0.094 & 0.725 & 4.355 & 0.906 & 53\\
            \hline
            6$\rightarrow$3   & 0.097 & 0.711 & 4.312 & 0.908 & 53 \\
            6$\rightarrow$6   & 0.087 & 0.698 & 4.234 & 0.914 & 68 \\
            \hline
            12$\rightarrow$3  & 0.098 & 0.73 & 4.370 & 0.900 & 53 \\
            12$\rightarrow$6  & 0.093 & 0.695 & 4.310 & 0.906 & 68 \\
            12$\rightarrow$12 & 0.089 & 0.692 & 4.242 & 0.910 & 99 \\
            
            \Xhline{4\arrayrulewidth}
        \end{tabular}
        \end{adjustbox}
    \caption{Ablation experiment for the DEQ iterations on the KITTI~\cite{geiger2012we} Eigen split. $a\rightarrow b$ represents $a :$ \# iters at training and $b :$ \# iters at test time.}
    \vspace{-7mm}
    \label{table:abldeq}
    \end{center}
\end{table}

            
\begin{table*}[t]
    \begin{center}
        \begin{adjustbox}{width=0.9\textwidth}
        \setlength{\tabcolsep}{10pt}
        \begin{tabular}{c|ccc|ccc}
            \Xhline{4\arrayrulewidth}
            \multirow{2}{*}{Methods} & \multicolumn{3}{c|}{Seq 9} & \multicolumn{3}{c}{Seq 10} \\
            \cline{2-7}
             & $t_{err}(\%)$ $\downarrow$ & $r_{err}(\degree/100m)$ $\downarrow$ & ATE ($m$) $\downarrow$ & $t_{err}(\%)$ $\downarrow$ & $r_{err}(\degree/100m)$ $\downarrow$ & ATE ($m$) $\downarrow$ \\
            \Xhline{4\arrayrulewidth}
            ORB-SLAM2~\cite{mur2017orb} (w/o LC) & 9.67 & 0.3 & 44.10 & 4.04 & 0.3 & 6.43 \\
            ORB-SLAM2~\cite{mur2017orb} & 3.22 & 0.4 & 8.84 & 4.25 & 0.3 & 8.51 \\
            \hline
            SfMLearner \cite{zhou2017unsupervised} & 19.15 & 6.82 & 77.79 & 40.40 & 17.69 & 67.34 \\
            GeoNet \cite{yin2018geonet} & 28.72 & 9.8 & 158.45 & 23.90 & 9.0 & 43.04 \\
            DeepMatchVO \cite{shen2019beyond} & 9.91 & 3.8 & 27.08 & 12.18 & 5.9 & 24.44 \\
            Monodepth2 \cite{godard2019digging} & 17.17 & 3.85 & 76.22 & 11.68 & 5.31 & 20.35 \\
            DW~\cite{gordon2019depth}-Learned & - & - & 20.91 & - & - & 17.88 \\
            DW~\cite{gordon2019depth}-Corrected & - & - & 19.01 & - & - & 14.85 \\
            SC-Depth~\cite{bian2019unsupervised} & 7.31 & 3.05 & 23.56 & 7.79 & 4.90 & 12.00 \\
            Zou \etal~\cite{zou2020learning} & \underbar{3.49} & \textbf{1.00} & \underbar{11.30} & \textbf{5.81} & \underbar{1.8} & 11.80 \\
            P-RGBD SLAM~\cite{bian2021unsupervised} & 5.08 & 1.05 & 13.40 & 4.32 & 2.34 & \textbf{7.99} \\
            \cellcolor{green!7} \textbf{DualRefine-initial ($T_0$)} & 9.06 & 2.59 & 39.31 & 9.45 & 4.05 & 15.13 \\
            \cellcolor{green!7} \textbf{DualRefine-refined ($T^*$)} & \textbf{3.43} & \underbar{1.04} & \textbf{5.18} & \underbar{6.80} & \textbf{1.13} & \underbar{10.85} \\
            \Xhline{4\arrayrulewidth}
        \end{tabular}
        \end{adjustbox}
    \caption{Results on Seq. 09 and Seq. 10 of the KITTI odometry data. We provide a comparison with other state-of-the-art self-supervised depth and odometry methods. ORB-SLAM2 is included as a representative non-learning based method. \textbf{Bold}: Best, \underbar{Underscore}: Second best. }
    \vspace{-7mm}
    \label{table:odom}
    \end{center}
\end{table*}
\begin{figure}[t!]
\begin{center}
    \begin{adjustbox}{width=0.4\textwidth}
    \setlength{\tabcolsep}{1pt}
    \begin{tabular}{c}
    {\tiny (a) Sequence 09} \\
    \includegraphics[width=0.25\linewidth]{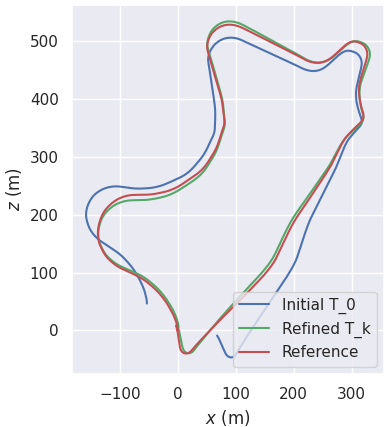} \\
    {\tiny (b) Sequence 10} \\
    \includegraphics[width=0.4\linewidth]{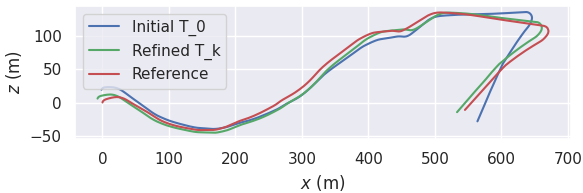} 
    \end{tabular}
    \end{adjustbox}
\caption{Estimated trajectory by the initial pose estimator and the refined trajectory using our pose refinement module on (a) Seq. 09 and (b) Seq. 10 of KITTI odometry data. The refined pose estimate improves the global trajectory, even without explicitly training for global consistency.}
\vspace{-7mm}
\label{fig:odom}
\end{center}
\end{figure}
\subsection{Depth results}
\tabref{table:comparison} shows the comparison of our depth estimation with SoTA self-supervised models. We compare with models that train on monocular video. Our model outperforms most previous models and is competitive with the Transformer~\cite{vaswani2017attention}-based DepthFormer~\cite{guizilini2022multi} model. Specifically, our model shows a significant improvement in $\delta_1$, suggesting highly accurate inliers. Furthermore, compared to DepthFormer that requires $~16$ GB of training memory per batch, ours only consumes $24$ GB of memory for 12 batches, around $1/8\times$. This is mainly because our method refines 2D hidden states based on local sampling, while \cite{guizilini2022multi} refines 2D feature maps \emph{and a 3D feature volume} using self\slash cross-attention along every depth bin. 

\figref{fig:res1} displays qualitative outputs for the disparity and error map of our model. We can observe an improvement to the error map of the refined depth. We also additionally display confidence weight outputs obtained by the model that computes them. We observe that the confidence $W_q$, which is calculated once, assigns the high confidence sparsely. Interestingly, the confidence weights that evolve with each iteration initially assign high confidence to far-away points and move towards closer points with increasing iterations.

\noindent\textbf{Limitation:} we note an increase in error for the moving vehicle in the lower image set. The worse RMSE of our model compared to DepthFormer also indicates higher outlier predictions. This could be due to repeated operation of our iterative updates, which may further exacerbate the outliers. Even with consistency masking, the model we propose displays limitations with dynamic objects. 
We leave further discussion of this issue for future study.

\subsection{Odometry results}
We present the results for visual odometry of the teacher model and the refinement model in \tabref{table:odom}. We also present the results of previous models that were trained on monocular videos. Our refinement module drastically improves the initial odometry results, as shown in \figref{fig:odom}. Although the goal of our study was to improve the estimation of local poses for accurate matching, we outperformed most of the other models in all metrics. Even without explicit training to ensure scale consistency, as in \cite{bian2019unsupervised, bian2021unsupervised}, our refined output demonstrates a globally consistent odometry prediction. Additionally, unlike Zou \etal~\cite{zou2020learning} which infers pose from long-term geometry, this result is achieved with only two input frames to infer pose estimates. Our model also achieves an ATE that is on par with traditional ORB-SLAM2, which performs global geometric optimization, although our results in $r_{err}$ still lag behind.

\label{sec:exp}

\section{Conclusions}

In this paper, we introduced a self-supervised pipeline for multi-frame depth and pose estimation and refinement. By leveraging the combined power of neural network representation and geometric constraints to refine both depth and pose, our approach achieved state-of-the-art performance in both tasks. Our method also demonstrates greater efficiency than competing methods, with potential for further improvement. Nevertheless, we still observed poorer depth accuracy in dynamic scenes. 

{\small
\bibliographystyle{ieee_fullname}
\bibliography{egbib}
}




\null\newpage
\null\newpage


{\huge{\textbf{Supplementary}}}
\vspace{10mm}


\section{DEQ Framework}
We adhere to the general framework of DEQ~\cite{bai2019deep, bai2022deep} and employ a quasi-Newton solver to accelerate convergence. In our experiments, we utilize the Anderson solver~\cite{anderson1965iterative}.
A DEQ model computes $A=I-\frac{\partial \mathrm{U}}{\partial z^*}$ at the fixed point $z^*$ to obtain the gradient. This is typically achieved by performing another fixed-point iteration. However, in line with \cite{fung2021fixed, geng2021attention, bai2022deep}, we approximate $A=I$ and utilize the inexact gradient for training.

\begin{figure}[h]
\begin{center}
    \includegraphics[width=0.45\textwidth]{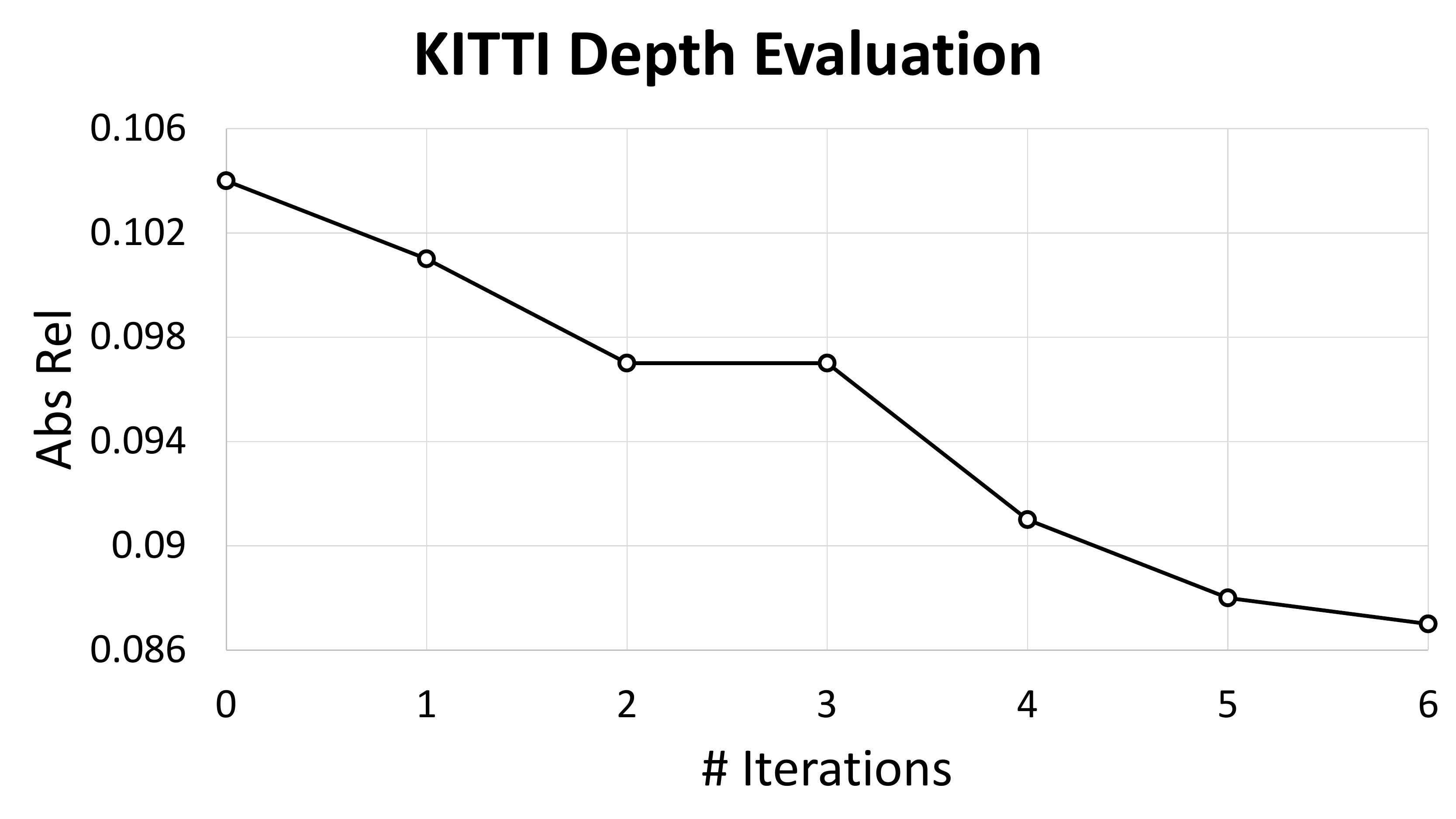} 
\caption{The progression of Abs Rel errors in each DualRefine iteration for KITTI depth.}
\vspace{-7mm}
\label{fig:iters}
\end{center}
\end{figure}

\begin{figure}[h]
\begin{center}
    \includegraphics[width=0.45\textwidth]{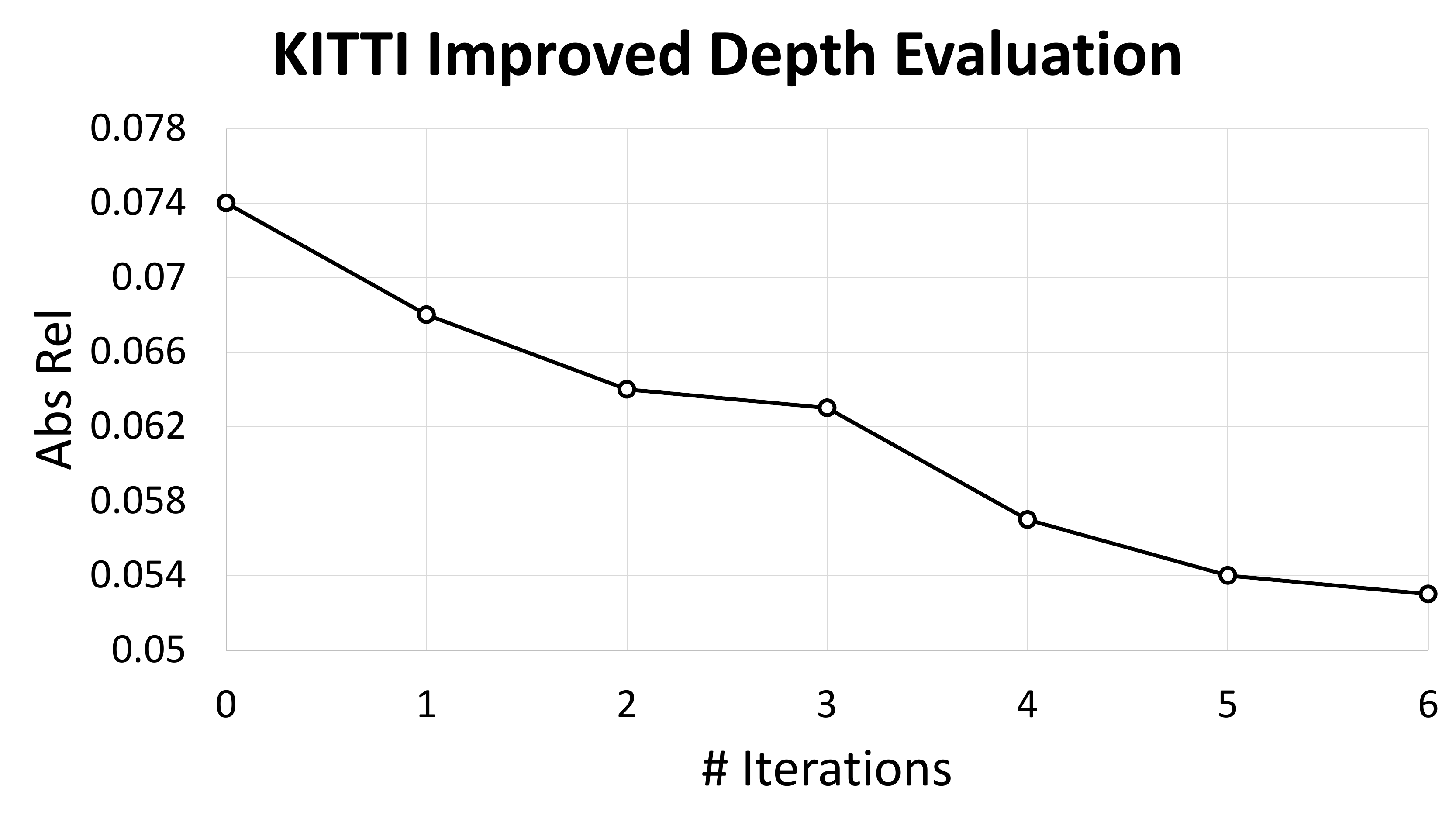}
\caption{The progression of Abs Rel errors in each DualRefine iteration for KITTI improved depth.}
\vspace{-7mm}
\label{fig:improvediters}
\end{center}
\end{figure}
\begin{table}[h]
        \begin{adjustbox}{width=0.48\textwidth}
        \begin{tabular}{c|cc|cccc}
            \Xhline{4\arrayrulewidth}
            &\multicolumn{2}{c|}{Loss pairs} & \multirow{2}{*}{Abs Rel $\downarrow$} & \multirow{2}{*}{Sq Rel $\downarrow$} &  \multirow{2}{*}{RMSE $\downarrow$} & \multirow{2}{*}{$\delta_1$ $\uparrow$} \\
            &$D^*$ & $T^*$ & & & & \\
            \Xhline{4\arrayrulewidth}
            1 & $T_0$ & $D_0$ & 0.99 & 0.765 & 4.449 & 0.898 \\
            2 & $T^*$ & $D_0$ & 0.093 & 0.698 & 4.342 & 0.907 \\
            3 & $T_0$ & $D^*$ & 0.092 & 0.657 & 4.34 & 0.908 \\
            4 & $T^*$ & $D^*$ & 0.089 & 0.632 & 4.305 & 0.907 \\
            \Xhline{4\arrayrulewidth}
        \end{tabular}
        \end{adjustbox}
    \caption{Ablation experiment for the effect of pose updates and self-supervision pairings on the KITTI~\cite{geiger2012we} Eigen split.}
    \label{table:ablpair}
\end{table}

\begin{table*}[t!]
    \begin{center}
        \begin{adjustbox}{width=0.95\textwidth}
        \begin{tabular}{cc|cc|ccccccc}
            \Xhline{4\arrayrulewidth}
            &Method & Test frames & $W\times H$ & \cellcolor{red!7} Abs Rel $\downarrow$ & \cellcolor{red!7} Sq Rel $\downarrow$ &  \cellcolor{red!7} RMSE $\downarrow$ & \cellcolor{red!7} RMSE log $\downarrow$ & \cellcolor{blue!7} $\delta_1$ $\uparrow$ & \cellcolor{blue!7} $\delta_2$ $\uparrow$ & \cellcolor{blue!7} $\delta_3$ $\uparrow$ \\
            \hline
            \multirow{11}{*}{ \rotatebox{90}{Low \& mid res}} &
            Ranjan~\cite{ranjan2019competitive} & 1 & $832\times 256$ & 0.123 & 0.881 & 4.834 & 0.181 & 0.860 & 0.959 & 0.985 \\
            &EPC++~\cite{luo2019every} & 1 & $832\times 256$ & 0.120 & 0.789 & 4.755 & 0.177 & 0.856 & 0.961 & 0.987 \\
            &Johnston~\cite{johnston2020self} \etal & 1 & $640\times 192$ & 0.081 & 0.484 & 3.716 & 0.126 & 0.927 & 0.985 & 0.996 \\
            &Monodepth2~\cite{godard2019digging} & 1 & $640\times 192$ & 0.090 & 0.545 & 3.942 & 0.137 & 0.914 & 0.983 & 0.995 \\
            &PackNet-SFM~\cite{guizilini20203d} & 1 & $640\times 192$ & 0.078 & 0.420 & 3.485 & 0.121 & 0.931 & 0.986 & 0.996 \\
            &\cellcolor{green!7}
            \textbf{DualRefine-initial ($D_0$)} & 1 & $640\times 192$ &  0.075	& 0.379	& 3.490	& 0.117	& 0.936	& 0.989	& \underbar{0.997} \\
            \arrayrulecolor{gray}\cline{2-11}
            &Patil \etal~\cite{patil2020don} & N$^\dagger$ & $640\times 192$ & 0.087 & 0.495 & 3.775 & 0.133 & 0.917 & 0.983 & 0.995 \\
            &Wang \etal~\cite{wang2020self} & 2 (-1, 0) & $640\times 192$ & 0.082 & 0.462 & 3.739 & 0.127 & 0.923 & 0.984 & 0.996 \\
            &ManyDepth~\cite{watson2021temporal} & 2 (-1, 0) & $640\times 192$ & 0.064 & 0.320 & 3.187 & 0.104 & 0.946 & 0.990 & 0.995 \\
            &DepthFormer~\cite{guizilini2022multi} & 2 (-1, 0) & $640\times 192$ & \underbar{0.055} & \textbf{0.271} & \textbf{2.917} & \underbar{0.095} & \underbar{0.955} & \underbar{0.991} & \textbf{0.998} \\
            &\cellcolor{green!7}
            \textbf{DualRefine-refined ($D^*$)} & 2 (-1, 0) & $640\times 192$ & \textbf{0.053} & \underbar{0.290} & \underbar{2.974} & \textbf{0.092} & \textbf{0.962} & \textbf{0.992} & \textbf{0.998} \\
            \Xhline{4\arrayrulewidth}
            \multirow{3}{*}{ \rotatebox{90}{High res}} &
            DRO~\cite{gu2021dro} & 2 (-1, 0) & $960\times 320$ & \underbar{0.057} & \underbar{0.342} & 3.201 & 0.123 & 0.952 & 0.989 & 0.996 \\
            &ManyDepth (HR ResNet50)~\cite{watson2021temporal} & 2 (-1, 0) & $1024\times 320$ & 0.062 & 0.343 & \underbar{3.139} & \underbar{0.102} & \underbar{0.953} & \underbar{0.991} & 0.997 \\
            &\cellcolor{green!7}
            \textbf{DualRefine-refined ($D^*$)} & 2 (-1, 0) & $960\times 288$ & \textbf{0.050} & \textbf{0.250} & \textbf{2.763} & \textbf{0.087} & \textbf{0.965} & \textbf{0.993} & \textbf{0.998} \\
            \Xhline{4\arrayrulewidth}
        \end{tabular}
        \end{adjustbox}
    \caption{Results and comparison with other state-of-the-arts models on the KITTI~\cite{geiger2012we} Eigen split~\cite{eigen2014depth} with improved depth maps~\cite{uhrig2017sparsity}. \textbf{Bold}: Best, \underbar{Underscore}: Second best. $^\dagger:$ evaluated on whole sequences}
    \vspace{-5mm}
    \label{table:improved}
    \end{center}
\end{table*}

\begin{table*}[h]
    \begin{center}
        \begin{adjustbox}{width=0.63\textwidth}
        \begin{tabular}{c|ccccccc}
            \Xhline{4\arrayrulewidth}
            $\#$ iters & \cellcolor{red!7} Abs Rel $\downarrow$ & \cellcolor{red!7} Sq Rel $\downarrow$ &  \cellcolor{red!7} RMSE $\downarrow$ & \cellcolor{red!7} RMSE log $\downarrow$ & \cellcolor{blue!7} $\delta_1$ $\uparrow$ & \cellcolor{blue!7} $\delta_2$ $\uparrow$ & \cellcolor{blue!7} $\delta_3$ $\uparrow$ \\
            \hline
             
             0 & 0.104 &	0.778 &	4.495 &	0.181 &	0.894 &	0.965 &	0.983 \\
             1 & 0.101 &	0.743 &	4.405 &	0.179 &	0.902 &	0.966 &	0.983 \\
             2 & 0.097 &	0.708 &	4.302 &	0.176 &	0.909 &	0.967 &	0.983 \\
             3 & 0.097 &	0.711 &	4.312 &	0.176 &	0.908 &	0.967 &	0.983 \\
             4 & 0.091 &	0.700 &	4.259 &	0.172 &	0.913 &	0.967 &	0.983 \\
             5 & 0.088 &	0.697 &	4.239 &	0.170 &	0.914 &	0.967 &	0.983 \\
             6 & 0.087 &	0.698 &	4.234 &	0.170 &	0.914 &	0.967 &	0.983 \\
             7 & 0.088 &	0.696 &	4.230 &	0.171 &	0.913 &	0.967 &	0.983 \\
             8 & 0.088 &	0.695 &	4.229 &	0.172 &	0.912 &	0.966 &	0.983 \\
             9 & 0.089 &	0.693 &	4.234 &	0.173 &	0.911 &	0.966 &	0.983 \\

            \Xhline{4\arrayrulewidth}
        \end{tabular}
        \end{adjustbox}
    \caption{The progression of the errors on the KITTI~\cite{geiger2012we} Eigen split in each DualRefine iteration.}
    \vspace{-5mm}
    \label{table:iters}
    \end{center}
\end{table*}

\begin{table*}[h!]
    \begin{center}
        \begin{adjustbox}{width=0.63\textwidth}
        \begin{tabular}{c|ccccccc}
            \Xhline{4\arrayrulewidth}
            $\#$ iters & \cellcolor{red!7} Abs Rel $\downarrow$ & \cellcolor{red!7} Sq Rel $\downarrow$ &  \cellcolor{red!7} RMSE $\downarrow$ & \cellcolor{red!7} RMSE log $\downarrow$ & \cellcolor{blue!7} $\delta_1$ $\uparrow$ & \cellcolor{blue!7} $\delta_2$ $\uparrow$ & \cellcolor{blue!7} $\delta_3$ $\uparrow$ \\
            \hline

             0 & 0.074 &	0.389 &	3.390 &	0.115 &	0.940 &	0.990 &	0.997 \\
             1 & 0.068 &	0.344 &	3.201 &	0.106 &	0.950 &	0.991 &	0.997 \\
             2 & 0.064 &	0.311 &	3.100 &	0.101 &	0.956 &	0.992 &	0.998 \\
             3 & 0.063 &	0.314 &	3.105 &	0.101 &	0.956 &	0.992 &	0.998 \\
             4 & 0.057 &	0.299 &	3.029 &	0.096 &	0.960 &	0.992 &	0.998 \\
             5 & 0.054 &	0.293 &	2.995 &	0.093 &	0.961 &	0.992 &	0.998 \\
             6 & 0.053 &	0.290 &	2.974 &	0.092 &	0.962 &	0.992 &	0.998 \\
             7 & 0.052 &	0.287 &	2.962 &	0.092 &	0.962 &	0.992 &	0.998 \\
             8 & 0.053 &	0.285 &	2.963 &	0.093 &	0.961 &	0.992 &	0.998 \\
             9 & 0.054 &	0.286 &	2.979 &	0.094 &	0.960 &	0.992 &	0.998 \\

            \Xhline{4\arrayrulewidth}
        \end{tabular}
        \end{adjustbox}
    \caption{The progression of the errors on the KITTI~\cite{geiger2012we} Eigen split~\cite{eigen2014depth} with improved depth maps~\cite{uhrig2017sparsity} in each DualRefine iteration.}
    \vspace{-5mm}
    \label{table:improvediters}
    \end{center}
\end{table*}

\section{Training Loss Combinations}
Determining the optimal pairings to calculate the self-supervision losses at the refined fixed point is not straightforward. For each refined estimate ($D^*$ and $T^*$), we can calculate the self-supervision loss using either the detached initial estimates ($[D^*\xleftrightarrow{}$ detached $T_0]$ pair and $[T^*\xleftrightarrow{}$ detached $D_0]$ pair) or the corresponding refined estimate ($[D^*\xleftrightarrow{}T^*]$ pair and $[D^*\xleftrightarrow{}T^*]$). 

The comparison is presented in \cref{table:ablpair}.
We observe a worse accuracy when both final estimates are paired with the corresponding initial estimates. We infer that, by pairing the final estimates with the initial ones, we impose a strong constraint on the model, limiting the scope of the output. 
We observe the best results when at least one of the final estimates is paired with the corresponding initial estimate. An example is when the depth loss is computed using the $[D^*\xleftrightarrow{}$ detached $T_0]$ pair, while the pose loss is computed using the $[T^*\xleftrightarrow{}D^*]$ pair. From this experiment, pairing the refined estimates with each other seems to display the best accuracy. 


\section{Additional results on KITTI Depth}

\subsection{KITTI improved depth}
In \cref{table:improved} we present evaluation results on improved dense ground truth~\cite{uhrig2017sparsity} of the KITTI~\cite{geiger2012we} eigen split~\cite{eigen2014depth}. We perform garg cropping~\cite{garg2016unsupervised} and report the error for distances up to 80$m$. Our refinement module improves the initial estimates and outperforms most previous models while still being competitive with the Transformer~\cite{vaswani2017attention}-based DepthFormer~\cite{guizilini2022multi} model.

\subsection{DEQ results}
In \cref{table:iters} we present the error for the output of our model in each DEQ iteration. Iteration $0$ corresponds to the depth estimates produced by the initial depth estimator. 
We can see that our model converges around the $6^{th}$ iteration. We also plot the Abs Rel error on \cref{fig:iters}

\subsection{KITTI improved depth DEQ results}
We also present detailed DEQ errors in \cref{table:improvediters} and plot the Abs Rel error in each iteration on \cref{fig:improvediters} for the KITTI improved depth ground truth. Similarly, our model converges around the $6^{th}$ iteration.

\subsection{Additional qualitative results}
We illustrate through \cref{fig:qualsupp,fig:episupp} additional results in the KITTI dataset. An interesting observation is how the model learns to give low confidence to vehicles and texture-less image regions. We also show in \cref{fig:episupp} how the epipolar geometry differs between the initial estimates and the refined estimates, which may cause inaccurate photometric losses as well as matching costs.

\null\newpage
\null\newpage
\section{Additional results on KITTI odometry}
\begin{table}[t]
    \begin{center}
        \begin{adjustbox}{width=0.5\textwidth}
        \setlength{\tabcolsep}{10pt}
        \begin{tabular}{c|ccc}
            \Xhline{4\arrayrulewidth}
            Methods & $t_{err}(\%)$ $\downarrow$ & $r_{err}(\degree/100m)$ $\downarrow$ & ATE ($m$) $\downarrow$  \\
            \Xhline{4\arrayrulewidth}
            ORB-SLAM2 [{\color{green}61}] & 12.96 & \textbf{0.7} & 44.09 \\
            \hline
            Monodepth2 [{\color{green}22}] & 12.28 & 3.1 & 99.36 \\
            Zou \etal [{\color{green}102}] & 7.28 & 1.4 & 71.63 \\
            \cellcolor{green!7} \textbf{DualRefine-initial ($T_0$)} & 12.50 & 4.04 & 118.29 \\
            \cellcolor{green!7} \textbf{DualRefine-refined ($T^*$)} & \textbf{5.82} & 1.51 & \textbf{17.27} \\
            \Xhline{4\arrayrulewidth}
        \end{tabular}
        \end{adjustbox}
    \vspace{-3.mm}
    \caption{Additional results on KITTI odometry test split (Seq. $11\sim 21$) using ORB-SLAM2 stereo as pseudo-GT. We provide a comparison with representative state-of-the-art self-supervised depth and odometry methods. ORB-SLAM2 is included as a representative non-learning based method. 
    }
    \vspace{-7.mm}
    \label{table:odomtest}
    \end{center}
\end{table}

We perform an additional evaluation on Seq. 11-21 of the KITTI odometry dataset, using the stereo version of ORB-SLAM2 as a pseudo-GT following Zou \etal~\cite{zou2020learning} We present the average results in \cref{table:odomtest} The refinement greatly improves over the initial predictions and also displays better ATE even in comparison to ORB-SLAM2 with loop closure.




\section{Conv-GRU Update Implementation}
In our approach, we use the standard Conv-GRU block~\cite{teed2020raft} to compute the updates as follows:
\begin{equation}
\begin{split}
    z_{k+1} = \sigma(\mathrm{CNN}_z([h_{k}, x_{k}])) \\
    r_{k+1} = \sigma(\mathrm{CNN}_r([h_{k}, x_{k}])) \\
    \tilde{h}_{k+1} = \mathrm{tanh}(\mathrm{CNN}_{\tilde{h}}([r_{k+1}\odot h_{k}, x_{k}])) \\
    h_{k+1} = (1-z_{k+1})\odot h_{k} + z_{k+1}\odot \tilde{h}_{k+1}
\end{split}
\label{eq:gru}
\end{equation}
where $\sigma$ represents the sigmoid activation function. Exploring other variants of the Conv-GRU block will be considered in the future.

\begin{figure*}[t!]
\begin{center}
    \begin{adjustbox}{width=0.95\textwidth}
    \setlength{\tabcolsep}{1pt}
    \begin{tabular}{c}
    \includegraphics[width=0.9\textwidth]{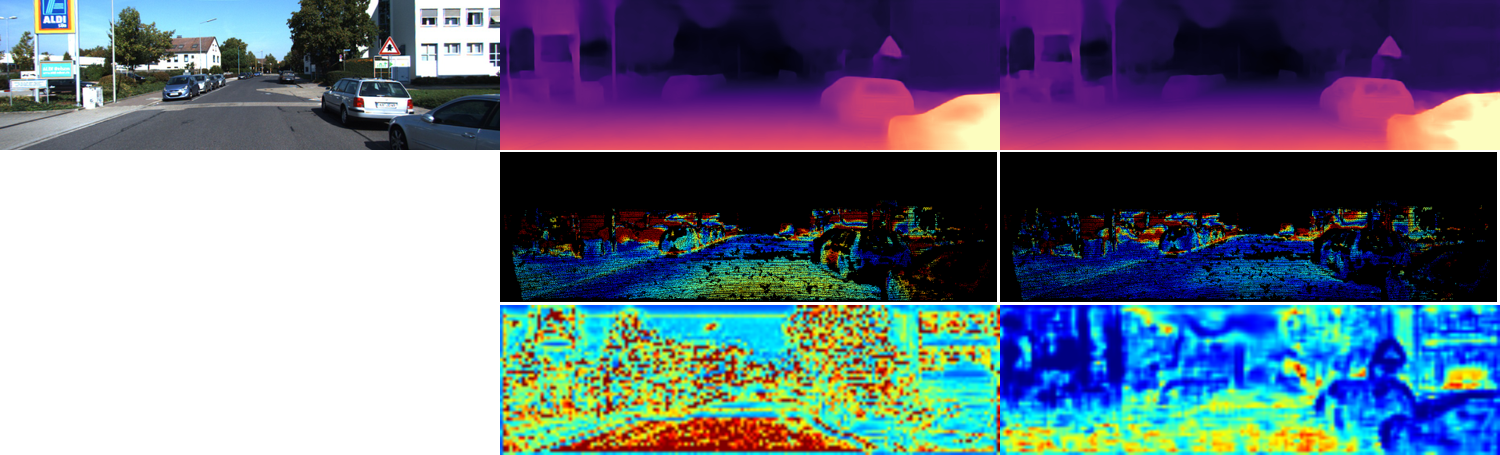} \\
            \Xhline{4\arrayrulewidth}
            \\
    \includegraphics[width=0.9\textwidth]{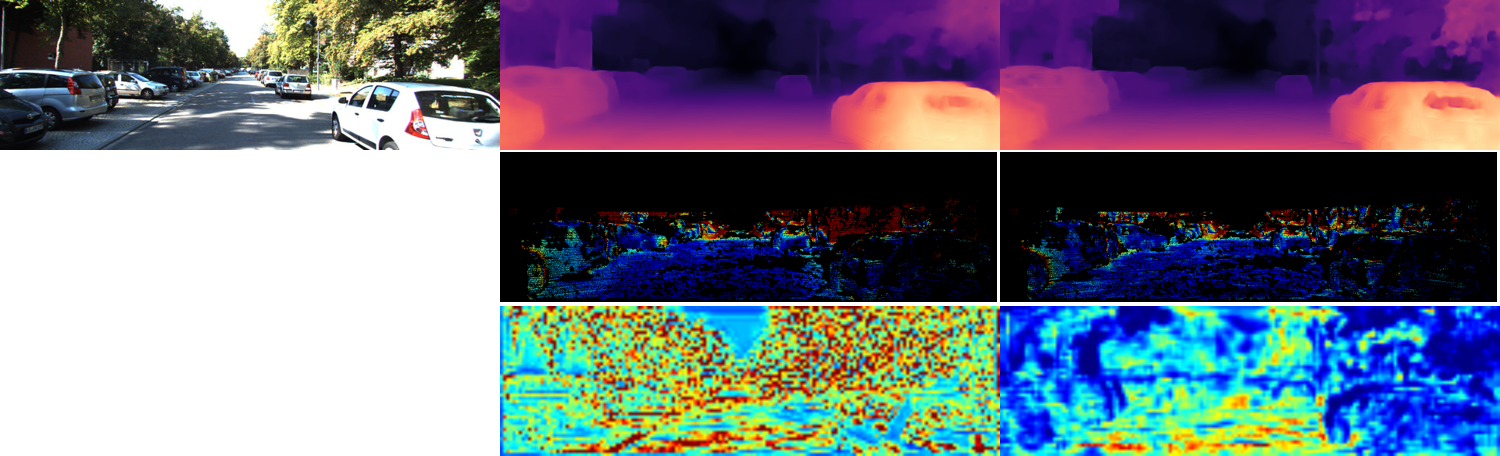} \\
            \Xhline{4\arrayrulewidth}
            \\
    \includegraphics[width=0.9\textwidth]{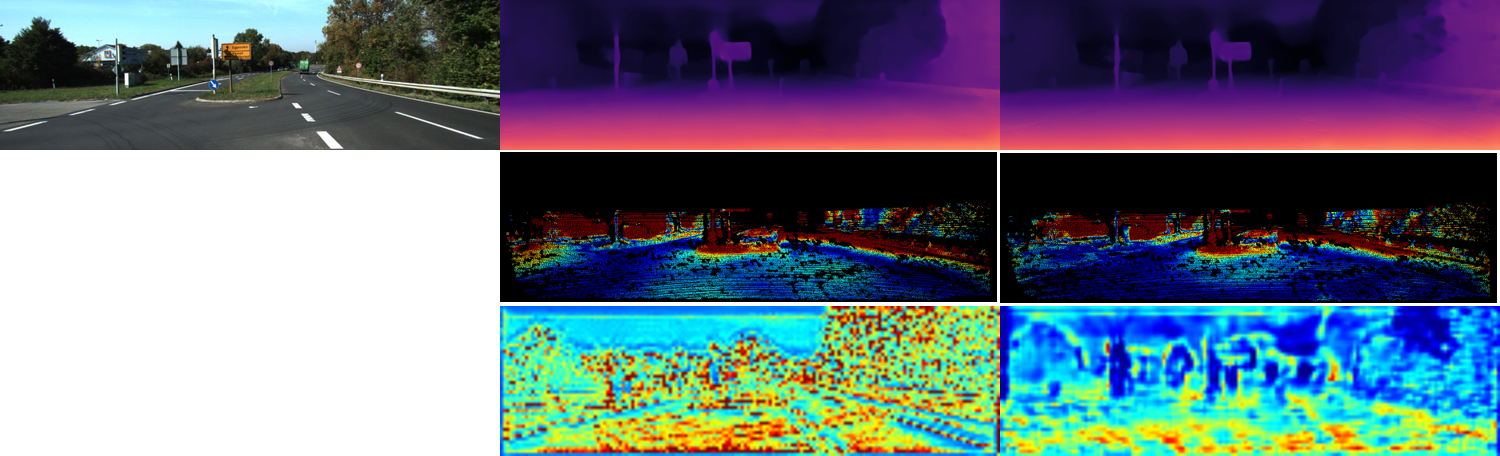} \\
            \Xhline{4\arrayrulewidth}
            \\
    \end{tabular}
    \end{adjustbox}
\caption{Qualitative results in the KITTI~\cite{geiger2012we} dataset. top left: input image, top center: initial disparity, top right: refined disparity, middle center: initial error map, middle right: refined error map, bottom center: fixed confidence weights, bottom right: $6^{th}$ iter confidence weights.}
\vspace{-5mm}
\label{fig:qualsupp}
\end{center}
\end{figure*}

\begin{figure*}[t!]
\begin{center}
    \begin{adjustbox}{width=0.75\textwidth}
    \setlength{\tabcolsep}{1pt}
    \begin{tabular}{c}
    \includegraphics[width=0.9\textwidth]{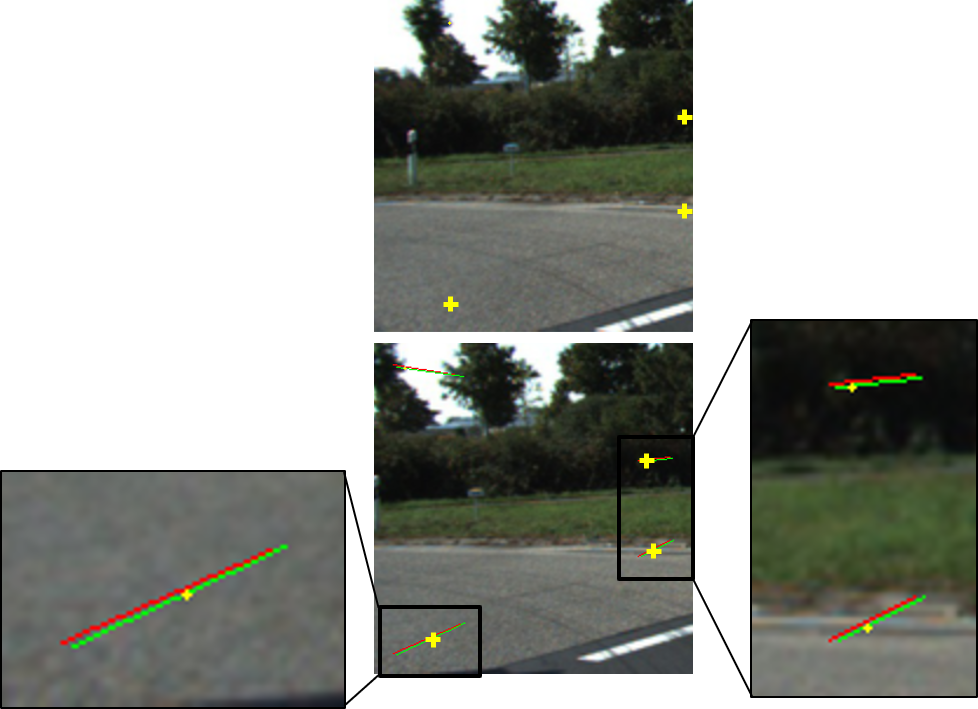} \\
            \Xhline{4\arrayrulewidth}
            \\
    \includegraphics[width=0.9\textwidth]{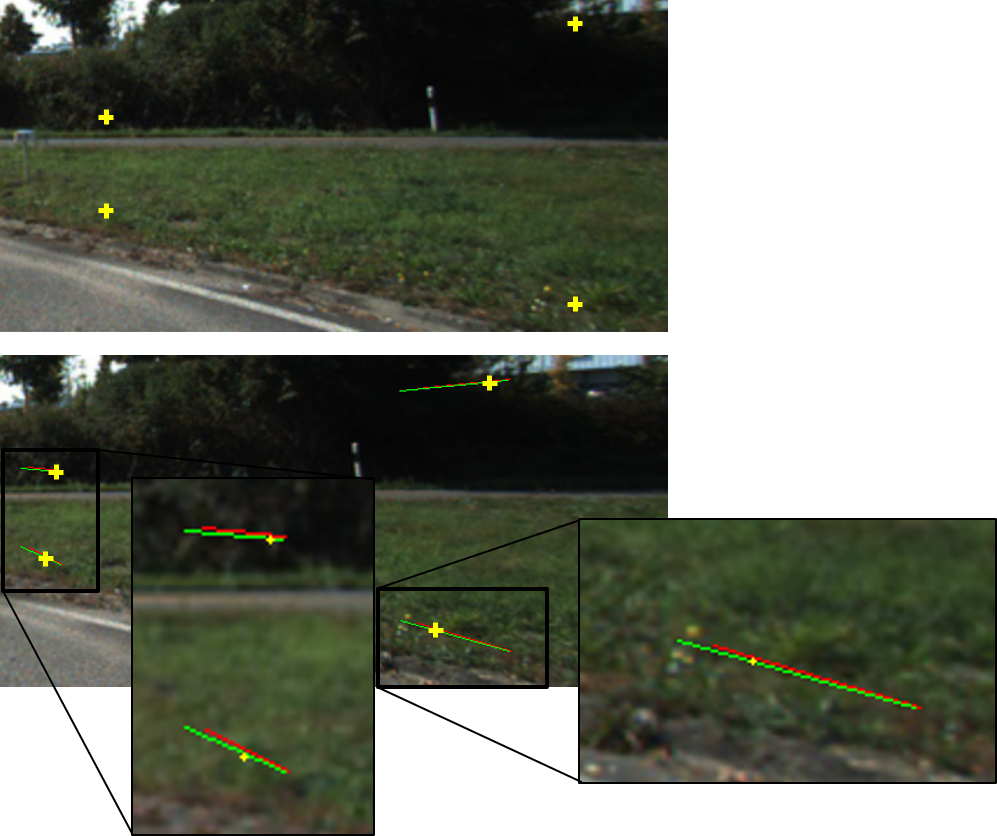} \\
    \end{tabular}
    \end{adjustbox}
\caption{The epipolar line in the source image, calculated from \textcolor{yellow}{yellow} points in the target image, for the PoseNet~\cite{kendall2015posenet} initial pose (\textcolor{red}{red}) and our refined pose (\textcolor{green}{green}). The yellow point in the source image is calculated based on our final depth and pose estimates.}
\vspace{-5mm}
\label{fig:episupp}
\end{center}
\end{figure*}




\end{document}